%% file: main.tex
\documentclass[conference]{IEEEtran}
\IEEEoverridecommandlockouts

\usepackage{cite}
\usepackage{amsmath,amssymb,amsfonts}
\usepackage{algorithmic}
\usepackage{graphicx}
\usepackage{textcomp}
\usepackage{xcolor}
\usepackage{subfigure}

\usepackage{tabularx, booktabs}
\usepackage{multirow}
\usepackage{array}
\usepackage{colortbl}
\definecolor{Gray}{gray}{0.95} 
\usepackage{graphicx}
\usepackage{amsmath}
\usepackage{amssymb}
\usepackage{booktabs}
\usepackage[pagebackref,breaklinks,colorlinks]{hyperref}
\input{preamble}
\usepackage{enumitem}
\usepackage[pagebackref,breaklinks,colorlinks]{hyperref}

\usepackage[capitalize]{cleveref}
\crefname{section}{Sec.}{Secs.}
\Crefname{section}{Section}{Sections}
\Crefname{table}{Table}{Tables}
\crefname{table}{Tab.}{Tabs.}

\begin{document}

\title{Scaling Prompt Instructed Zero Shot Composed Image Retrieval with Image-Only Data
}

\author{
	Yiqun Duan\textsuperscript{1,2}\thanks{This work was completed by Yiqun during his internship at Amazon, Australia, during August 2023 to November 2023. This work is published at IJCNN 2025.}  \ \ \ \
    Sameera Ramasinghe\textsuperscript{1}  \ \ \ \
    Stephen Gould\textsuperscript{2,3}  \ \ \ \
    Ajanthan Thalaiyasingam\textsuperscript{1}  \ \ \ \
    \\
	\textsuperscript{1} Amazon \ \ \ \
    \textsuperscript{2} University of Techonlogy Sydney  \ \ \ \ \textsuperscript{3} The Australian National University
	
}


\maketitle

\begin{abstract}
Composed Image Retrieval (CIR) is the task of retrieving images matching a reference image augmented with a text, where the text describes changes to the reference image in natural language. Traditionally, models designed for CIR have relied on triplet data containing a reference image, reformulation text, and a target image. However, curating such triplet data often necessitates human intervention, leading to prohibitive costs. This challenge has hindered the scalability of CIR model training even with the availability of abundant unlabeled data. With the recent advances in foundational models, we advocate a shift in the CIR training paradigm where human annotations can be efficiently replaced by large language models (LLMs). Specifically, we demonstrate the capability of large captioning and language models in efficiently generating data for CIR only relying on unannotated image collections. Additionally, we introduce an embedding reformulation architecture that effectively combines image and text modalities. Our model, named InstructCIR, outperforms state-of-the-art methods in zero-shot composed image retrieval on CIRR and FashionIQ datasets. Furthermore, we demonstrate that by increasing the amount of generated data, our zero-shot model gets closer to the performance of supervised baselines.
\end{abstract}



\input{sec/1_intro}

\input{sec/2_method}

\input{sec/4_relatedworks}

\input{sec/3_exp}

\input{sec/5_conclusion}


{\small
\bibliographystyle{ieee_fullname}
\bibliography{main}
}

\end{document}

%% file: preamble.tex
\newcommand{\e}[1]{\mathbf{e}#1}
\usepackage{arydshln}

\usepackage{xspace} 

\newcommand{\eg}{\textit{e.g.}\xspace}

\newcommand{\etc}{\textit{etc.}\xspace}

\newcommand{\etal}{et al.\xspace}
\usepackage{adjustbox}

\newcommand{\x}[1]{\mathbf{x}#1}

\definecolor{songye}{HTML}{42602D}
\definecolor{hong2}{HTML}{DB4D6D}
\definecolor{redmean}{HTML}{A12026}
\definecolor{greenmean}{HTML}{335D17}
\definecolor{redgraff}{HTML}{B1001C}
\definecolor{greengraff}{HTML}{235E00}
\definecolor{greycolor}{HTML}{877F6C}
\definecolor{amethyst}{rgb}{0.6, 0.4, 0.8}

%% file: sec/1_intro.tex
\section{Introduction}
\label{sec:intro}
The objective of composed image retrieval~\cite{hosseinzadeh2020composed,dodds2020modality_maaf} is to search for an image that aligns with both a reference image and a textual input detailing the desired alterations to that reference.  
This allows users to modify an image-based search query with natural language, facilitating a clear articulation of intent. Such capabilities have wide-ranging applications, including e-commerce, recommendation systems, and search engines.


The effectiveness of recent CIR methods~\cite{vo2019composing,dodds2020modality_maaf,chen2020image_val,liu2021image,Baldrati_2022_CVPR_clip4cir0} largely depends on the pre-trained vision and language models such as CLIP~\cite{radford2021learning} and BLIP~\cite{li2022blip}, which utilize contrastive semantic matching. Nonetheless, these models need to be finetuned specifically for CIR using triplet data containing reference images, reformulation texts, and target images. This reliance on human-annotated triplet data hinders the scalability of CIR models. Additionally, the scarcity of triplet data can impede fine-grained reformulations across the interrelated visual and text modalities.

\begin{figure}[hbpt]
    \includegraphics[width=1.02\columnwidth]{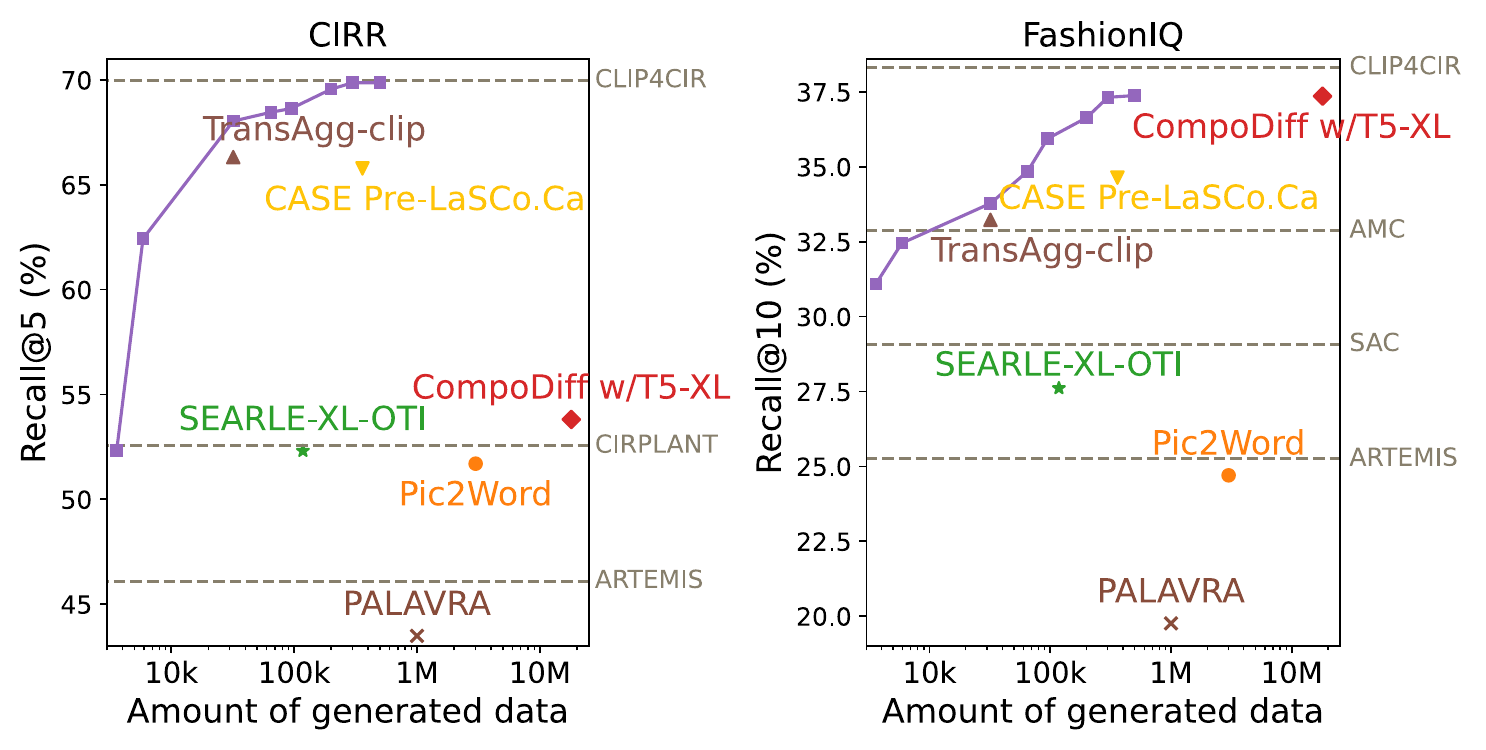}
    \vspace{-3ex}
    \caption{ \label{fig:cover} Performance curve versus current zero shot composed image retrieval benchmarks, where \textcolor{greycolor}{grey dashlines - -} indicates supervised baselines as intuitive references. Our zero-shot model (shown in \textcolor{amethyst}{purple}) closes the gap with supervised baselines with increasing amount of generated data.}
\end{figure}

In contrast, two parallel studies, Pic2Word~\cite{saito2023pic2word} and ZS-CIR~\cite{baldrati2023zero}, introduced the zero-shot composed image retrieval task by leveraging image inversion techniques~\cite{gal2022image}. Both methods convert the reference image into a single text token and apply reformulation in the text domain. However, this transformation of images to singular text tokens may result in substantial loss of intricate visual details, leading to sub-par zero-shot performance.

A concurrent work, TransAgg~\cite{liu2023zero}, suggests the use of an LLM (ChatGPT) to produce training data from existing image caption datasets. Yet, this approach remains reliant on pre-existing image-caption paired data and uses generically trained image and text encoders while only optimizing an aggregation layer. Moreover, TransAgg has been explored only on a limited data scale.


Our approach further relaxes the data requirement by solely relying on unannotated image collections. We hypothesize that given an arbitrary image pair, one can generate the natural language description of the difference by utilizing large vision and language models, effectively replacing human annotations.
To this end, we first leverage the LLaVa model~\cite{liu2023visual} to generate captions for a randomly sampled image pair. Subsequently, we utilize an LLM to delineate the differences between these captions. This technique enables the formation of triplets in a zero-shot manner, relying exclusively on unannotated image collections.

Additionally, we introduce a simple, yet effective joint embedding reformulation architecture that fuses the image and text modalities using cross-attention at multiple levels. Such a latent fusion design enables fine-grained image manipulations using text and has been used in text-guided image generation~\cite{rombach2022high,zhang2023adding,bubeck2023sparks}. Nevertheless, this embedding reformulation approach has not been used in the CIR domain to our knowledge and outperforms late fusion techniques as well as direct pixel-space manipulations as shown in our experiments. Our contributions are three-fold: 
\begin{itemize}[noitemsep,topsep=0pt,parsep=0pt,partopsep=0pt]
    \item We introduce a scalable framework for training CIR models that relies solely on unannotated image collections, replacing the need for paired image-caption data and human annotations.
    \item We introduce a simple, yet effective latent fusion architecture to effectively combine image and text modalities for CIR. 
    \item Our model sets the state-of-the-art on zero-shot CIR benchmarks and closes the gap between zero-shot and supervised settings aided by efficient data scaling; See Fig.~\ref{fig:cover}.
\end{itemize}

\begin{figure}[t]
\centering
    \includegraphics[width=0.95\columnwidth]{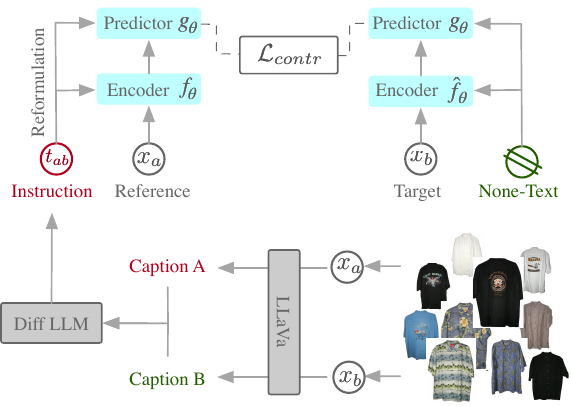}
    \caption{InstructCIR workflow. Given an image pair $\{\x_a, \x_b\}$, captions are generated by the LLaVA model. Next, the LLM generates the natural language description of the differences between these captions. Then, the reference image and the generated difference text are fused at multiple levels and the model is trained to minimize the embedding distance with respect to the target image. \label{fig:schema}}
\end{figure}

%% file: sec/2_method.tex
\section{InstructCIR}
This section presents a detailed overview of our zero-shot triplet data generation pipeline (Sec.~\ref{subsec:pipeline}) and model architecture (Sec.~\ref{subsec:reformulationnetwork}). Section~\ref{subsec:cirtrain} elaborates on the training process as well as objective functions for composed image retrieval.

\subsection{Zero-Shot Triplet Generation} ~\label{subsec:pipeline}
A comprehensive workflow of our approach is depicted in Fig.~\ref{fig:schema}. Initially, an image pair is randomly selected from the provided image collection. This pair is then transformed into textual descriptions via image captioning. Subsequently, the LLMs are prompted with these generated captions to produce the reformulation text that describes the difference between the captions. 

\begin{figure*}[t]
    \centering
    
    \raisebox{2cm}{
    \resizebox{0.98\textwidth}{!}{
    \begin{minipage}[b]{1.2\textwidth}
        \begin{tabularx}{1.0\linewidth}{cc cc}
        \toprule
        Image & Converted Caption & Image & Converted Caption \\ \midrule
        \multicolumn{4}{c}{Examples on FashionIQ Dataset} \\ \hdashline
        
            \includegraphics[width=0.06\linewidth]{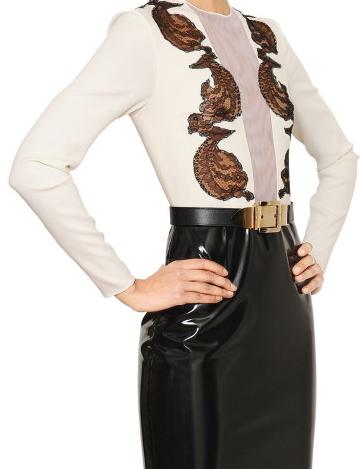} & 
            \raisebox{0.65cm}{
            \begin{minipage}{0.4\linewidth}
             \small{
                    The image features a woman wearing \textcolor{songye}{a white shirt and a black leather pencil skirt}. \textcolor{hong2}{She is posing for the camera, standing with her hands on her hips.} The woman is wearing \textcolor{songye}{a gold belt}, which adds a touch of \textcolor{songye}{elegance} to \textcolor{hong2}{her overall appearance.}}
                \end{minipage} }&
            \includegraphics[width=0.04\linewidth]{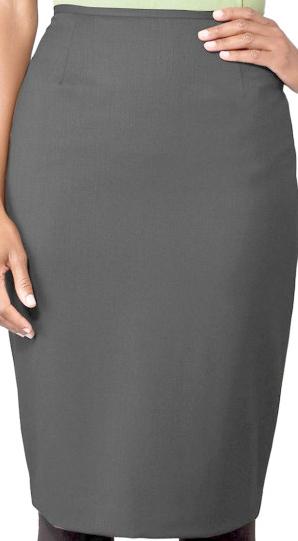} &  
            \raisebox{0.65cm}{
            \begin{minipage}{0.4\linewidth}
            \small{
                    The image features a woman wearing \textcolor{songye}{a gray skirt and a green shirt}.  The woman \textcolor{hong2}{is positioned in the center of the frame}, with her \textcolor{songye}{skirt covering most of her lower body}. The overall scene appears to be a \textcolor{songye}{casual and comfortable setting}.}
                \end{minipage} }\\ \hdashline
            \includegraphics[width=0.035\linewidth]{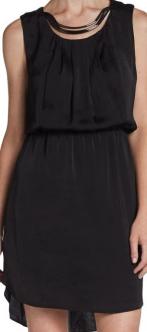} & \raisebox{0.8cm}{ \begin{minipage}{0.4\linewidth}
            \small{
                   The image features a woman wearing a \textcolor{songye}{black dress with a necklace around her neck}. She is \textcolor{hong2}{standing in front of a mirror, possibly posing for a picture}. The dress appears to be a \textcolor{songye}{short, black dress with a neckline that accentuates her necklace.}}
                \end{minipage}
                } &
            \includegraphics[width=0.03\linewidth]{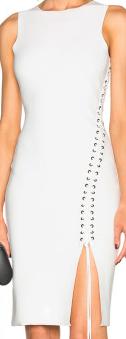} &   \raisebox{0.8cm}{\begin{minipage}{0.4\linewidth}
            \small{
                   The image features a woman wearing \textcolor{songye}{a white dress, which appears to be a strapless, \textcolor{hong2}{off-the-shoulder}, split style}. The dress has a unique design, \textcolor{songye}{with a row of buttons running down the front}. \textcolor{hong2}{She is holding a handbag, and the scene exudes elegance and sophistication. } The overall style of the dress is \textcolor{songye}{elegant}.}
                \end{minipage}}\\ \hdashline
            Reform & 
            \begin{minipage}{0.4\linewidth}
            \small{
                   Changed white shirt and skirt to short black dress, added necklace.}
                \end{minipage}
            & Reform& \begin{minipage}{0.4\linewidth}
            \small{
                   From gray skirt and green shirt to elegant white strapless split dress.}
                \end{minipage} \\ \midrule
                \end{tabularx}
                
                \begin{tabularx}{1.00\linewidth}{cc cc}
                \multicolumn{4}{c}{Examples on CIRR Dataset} \\ \hdashline
                \begin{adjustbox}{margin=-0.3cm 0cm 0cm 0cm}
                \includegraphics[width=0.06\linewidth]{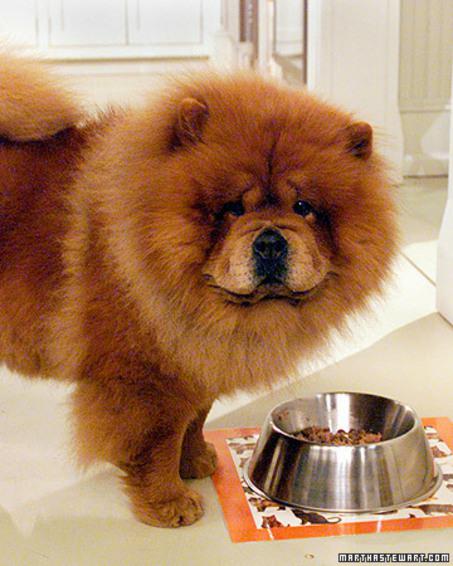}\end{adjustbox} & \raisebox{0.7cm}{
                \begin{adjustbox}{margin=-0.2cm 0cm 0cm 0cm}
                \begin{minipage}{0.4\linewidth}
            \small{
                   The image features \textcolor{songye}{a brown, shaggy dog standing} on \textcolor{hong2}{a dining table} \textcolor{songye}{next to a bowl of food}. The dog appears to be \textcolor{songye}{looking at the camera}, \textcolor{hong2}{possibly waiting for its owner to take a picture}. \textcolor{hong2}{The scene captures the dog's curiosity} as it stands near the bowl. }
                \end{minipage}
                \end{adjustbox}
                } &
                \begin{adjustbox}{margin=-0.1cm 0cm 0cm 0cm}
                \raisebox{0.35cm}{
            \includegraphics[width=0.08\linewidth]{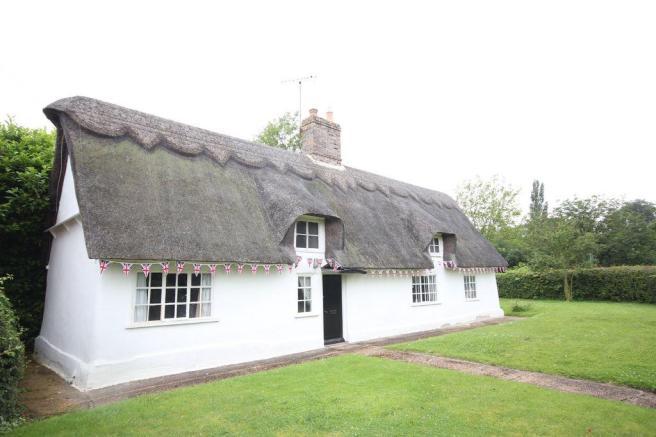} } \end{adjustbox}& 
            \begin{adjustbox}{margin=-0.34cm 0cm 0cm 0cm}\raisebox{0.65cm}{ \begin{minipage}{0.39\linewidth}
            \small{
                   The image features \textcolor{songye}{a white house with a thatched roof}, \textcolor{hong2}{giving it a charming and rustic appearance.} \textcolor{songye}{The house is situated on a lush green field, surrounded by a beautiful garden.} The \textcolor{hong2}{garden is adorned with a flag, adding a touch of patriotism} to the scene.  }
                \end{minipage}
                } \end{adjustbox} \\ \hdashline
                \begin{adjustbox}{margin=-0.3cm 0cm 0cm 0cm}
                \raisebox{0.35cm}{
                \includegraphics[width=0.07\linewidth]{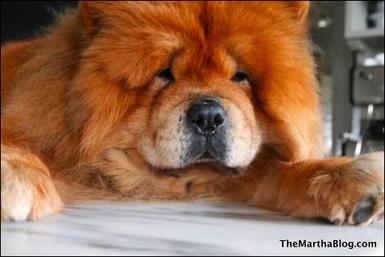}} \end{adjustbox} & 
                \begin{adjustbox}{margin=-0.25cm 0cm 0cm 0cm}
                \raisebox{0.9cm}{\begin{minipage}{0.4\linewidth}
            \small{
                  The image features \textcolor{songye}{a large, fluffy, and furry dog lying on a white countertop}. The dog appears to be a \textcolor{songye}{Chow Chow}, with its \textcolor{hong2}{distinctive appearance} and \textcolor{songye}{long, shaggy fur}. The dog is positioned in the center of the scene, \textcolor{songye}{occupying a significant portion of the countertop}.}
                \end{minipage}} \end{adjustbox} &
                \begin{adjustbox}{margin=-0.25cm 0cm 0cm 0cm}
                \raisebox{-0.1cm}{
            \includegraphics[width=0.045\linewidth]{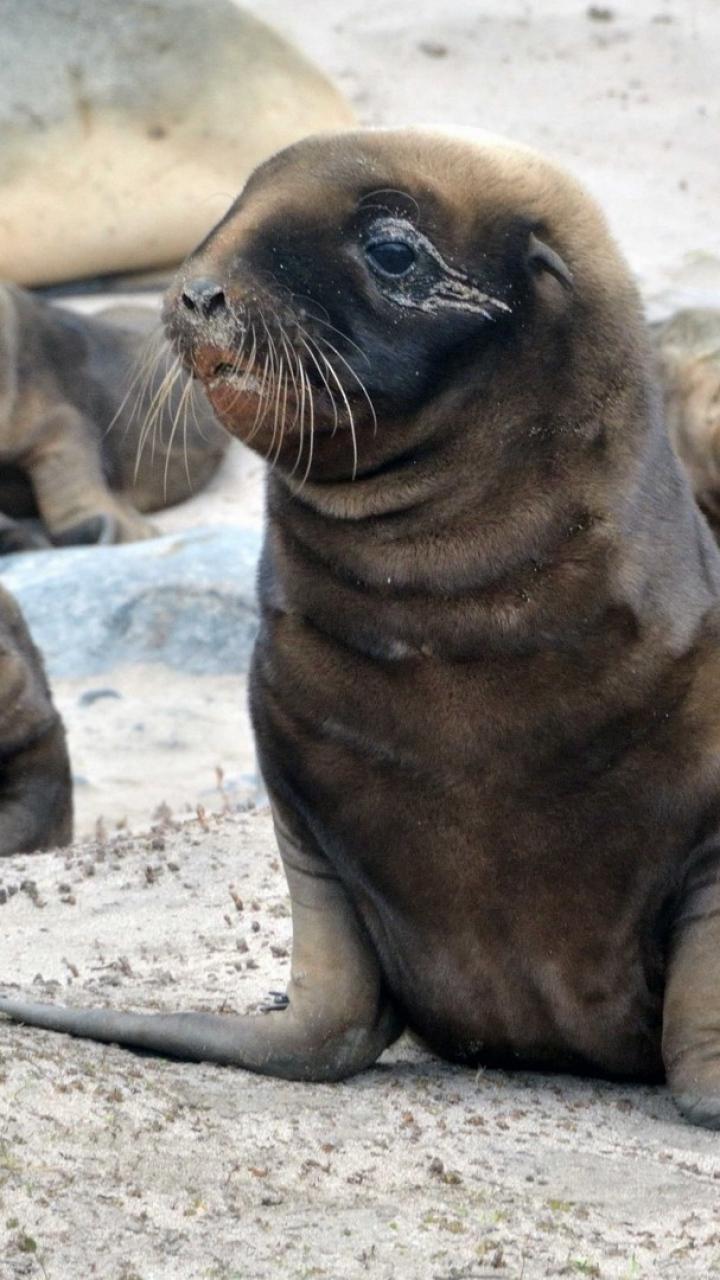} } \end{adjustbox}&  
             \begin{adjustbox}{margin=-0.25cm 0cm 0cm 0cm}\raisebox{0.8cm}{\begin{minipage}{0.39\linewidth}
            \small{
                  The image features \textcolor{songye}{a group of seals sitting on a sandy beach}. There are at least four seals visible in the scene, with \textcolor{songye}{one seal sitting prominently in the foreground and the others scattered around the beach}. The seals \textcolor{hong2}{appear to be enjoying their time} on the sand, possibly \textcolor{hong2}{resting or socializing} with each other.  }
                \end{minipage}} \end{adjustbox}\\ \hdashline
            Reform & 
            \begin{minipage}{0.4\linewidth}
            \small{
                   Dog moved from dining table to white countertop, now lying down.}
                \end{minipage}
            & Reform& \begin{minipage}{0.4\linewidth}
            \small{
                   House with garden and flag changes to seals on a sandy beach.}
                \end{minipage} \\ 
                \bottomrule
        \end{tabularx}
    \end{minipage}
    }
    }
    \caption{\label{fig:caption}Caption results by prompting visual LLMs on Fashion200k and NLVR, where \textcolor{songye}{accurate information} and \textcolor{hong2}{unrelated information} is highlighted in different color. Though unrelated information exists, it still suggests zero-shot image-text conversion is feasible with LLMs. Reform corresponds to the reformulation text generated for the images in the same column. }
\end{figure*}

\paragraph{Image Captioning:}
One key component in ensuring zero-shot CIR only depends on the given image collection is to convert images into a sufficiently detailed text description. 
We prompt a powerful visual instructed large language model (LLaVA\footnote{\href{https://huggingface.co/liuhaotian/llava-v1.5-13b}{https://huggingface.co/liuhaotian/llava-v1.5-13b}}~\cite{liu2023visual}) to convert the sampled image pairs into a text pair. 
The prompt for the captioning model is provided in the top part of Table~\ref{tb:reformulationprompts}, and example results are shown in Fig.~\ref{fig:caption}.
The image is tokenized and encoded through a ViT model and is fed into an LLM as context tokens. The LLM generates the detailed image caption given visual tokens and task prompts.
The examples on the Fashion200k dataset show that, although LLM may generate common descriptions such as ``posing for a photography shot", ``a woman is wearing", it still accurately describes key elements of the given image, such as the object (\eg., shirt, dress), style (\eg., strapless, off-the-shoulder, split style), color, \etc.

\begin{table}[t]
\caption{\label{tb:reformulationprompts} The Upper part denotes prompts used to convert images into texts. The lower part indicates text prompts used for reformulating from image caption A to caption B. }
\centering
\begin{minipage}[b]{0.43\textwidth}
        \includegraphics[width=\textwidth]{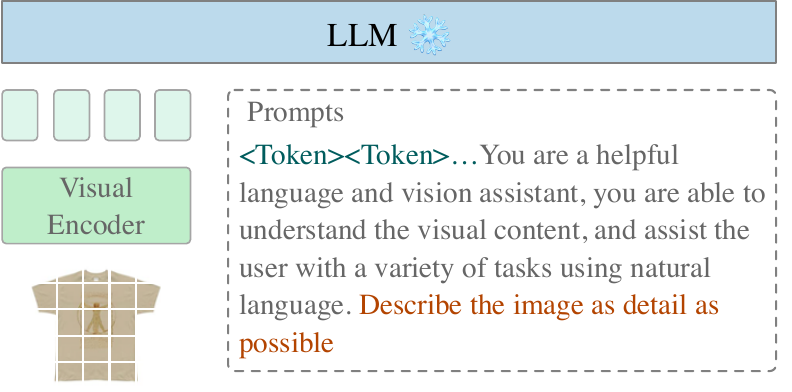}
    \end{minipage}
    \hfill
\begin{tabularx}{0.9\linewidth}{X}
\toprule
Reformulation Prompt: \\ \midrule
\small{
You have two captions for two images, image A and image B, you are supposed to write a reformulation text describing changing from image A to image B. 
caption A: \textcolor{redmean}{\{Caption A\} }
caption B: \textcolor{greenmean}{ \{ Caption B \}}
answer should be concise and within 12 words, only contain normal words, do not use special characters. 
Difference:} \\ \bottomrule
\end{tabularx}
\vspace{-8pt}
\end{table}

\paragraph{Reformulation with LLMs:}
Utilizing the high-quality image captions obtained as above, our approach leverages an LLM fine-tuned from vicuna 33B checkpoint\footnote{\href{https://huggingface.co/lmsys/vicuna-33b-v1.3}{https://huggingface.co/lmsys/vicuna-33b-v1.3}} to generate reformulated text-prompts \textbf{in a zero-shot manner.} 
The language model is fine-tuned
using low-rank adaptation fine-tuning (LORA)~\cite{hu2021lora} with 10 epochs. 
The fine-tuning seed data is obtained by prompting ChatGPT-4 in zero-shot. 
Here, directly prompting ChatGPT-4 is also feasible to generate reformulation data. Here we fine-tuning our own model for the feasibility of the ablation study. 
We employ the prompt structure illustrated in Table~\ref{tb:reformulationprompts}, which is crafted without prior training specific to this task.
Illustrative examples of generated reformulation texts are presented in Fig.~\ref{fig:caption}. These highlight the efficacy of our data generation pipeline. Rather than merely presenting experiment outcomes, we delve deeper to discern the influence of various language models on the performance metrics.

\begin{figure}[t]
    \centering
    \includegraphics[width=0.8\columnwidth]{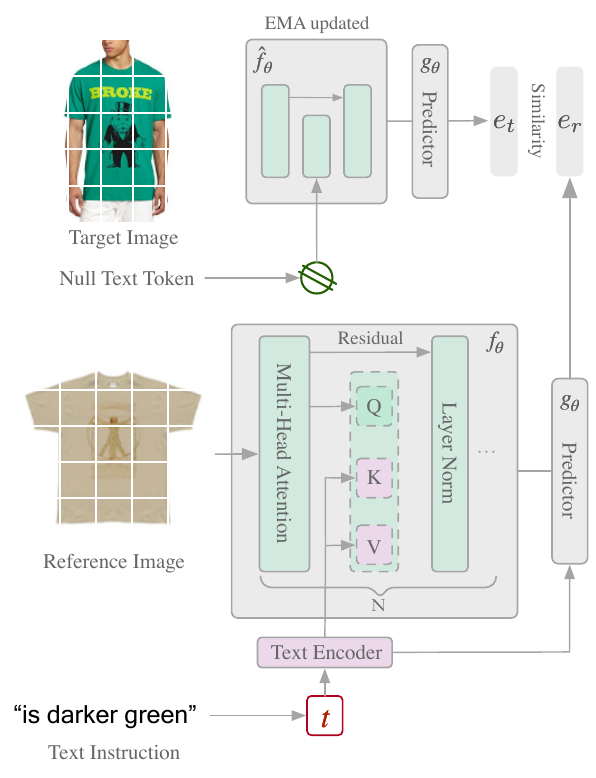}
    \caption{\label{fig:reformer} Model structure of proposed embedding reformulation network. The text embedding is injected into the visual transformer through cross-attention layer by layer. }
\end{figure}

\subsection{Model Structure} \label{subsec:reformulationnetwork}
 Our proposed architecture capitalizes on a streamlined yet potent framework; a layer-by-layer  \textbf{text-guided embedding reformulation network}, as illustrated in Fig.~\ref{fig:reformer}. 
This design is inspired by in the latest advancements in generative image manipulation~\cite{rombach2022high,brooks2022instructpix2pix} where models adeptly utilize text instructions to guide image generation. 
Although the generated images are visually plausible, they fail on recall metrics (as shown in Table~\ref{tb:zero-shotmain}) due to overly strict pixel-to-pixel correlations. 
Instead of the pixel space, 
we harness the joint latent space~\cite{assran2023self} to perform the instructed embedding reformulation.  

We modify each transformer block of the ViT model~\cite{dosovitskiy2021an} by introducing cross attention from the text encoder. 
Our text encoder is the same as CLIP~\cite{radford2021learning}, which takes text instruction as input and outputs an embedding sequence. 
In each ViT block, one self-attention layer is first applied on visual embeddings and cross-attention is utilized to inject the text instruction on embeddings, where the text embeddings are treated as $K$ and $V$ and image embeddings as queries $Q$.
Then, an MLP layer combined with layer norm~\cite{ba2016layer} and residual connection is utilized to aggregate the features.
Several such reformulation blocks are stacked to get the final output.
On the top of the encoder, a predictor is used to obtain the final embedding for retrieval. 
For the target images, which do not have an instruct prompt, we use a null-text token as the text input. 
This enables us to embed both the source and target images within a shared model architecture.
Additionally, the target encoder is updated using an exponential moving average similar to I-JEPA~\cite{assran2023self}.

\subsection{Composed Retrieval Training}~\label{subsec:cirtrain}
We train the joint-embedding reformulation network by minimizing the distance between the reformulated query embeddings and the target embeddings. For a fair comparison, we maintain the model size consistent with preceding models which are similar to the CLIP model.
For a given query with an image pair $\x_a, \x_b$, we obtain the associated captions $C_a, C_b$ using the LLaVa model. Subsequently, the reformulation LLM produces the reformulation text $t_{ab}$ in a zero-shot fashion, defined as $t_{ab} = \text{LMM}_{\text{diff}}(C_a, C_b)$. 

Subsequently, the reformulated embedding $\e_r$ is extracted by our proposed encoder and a predictor. Formally,
\begin{equation}
    \e_r = g_{\theta}(\mathbf{z}_r, \phi_t(t_{ab})), \quad\mbox{where $\mathbf{z}_r = f_{\theta}(\x_a, t_{ab})$},
\end{equation}
and $\phi_t(\cdot)$ is the pre-trained text encoder. 
The predictor $g_\theta(\cdot)$ is a MLP projection layer based on concatenated image and text embeddings $\{\mathbf{z}_r, \phi_t(t_{ab})\}$. 
Detailed description of the encoder $f_{\theta}(\cdot, \cdot)$ is provided in the model structure section (Sec.~\ref{subsec:reformulationnetwork} and Fig.~\ref{fig:reformer}).
The target embedding is extracted using the same function with the null-text token embedding instead of input text, $\e_t = g_{\theta}(\mathbf{z}_t, \phi_t(\emptyset))$, where $\mathbf{z}_t = f_{\theta}(\x_b, \emptyset)$, 
Here, $\e_r, \e_t \in \mathbb{R}^{d}$ where $d$ is embedding dimensionality.

Our training objective is to minimize the distance between the query embedding $\e_r$ and the target embedding $\e_t$ for a given triplet $\{\x_a, \x_b, t_{ab}\}$ from the zero-shot pipeline. Simultaneously, we maximize the distance between $\e_r$ and embeddings of other target images within the batch. 
To accomplish this, we utilize a batch-centric contrastive loss:

\begin{equation}
    \mathcal{L}_{\text{contr}} = \frac{1}{B} \sum^{B}_{i=1} -\text{log} \frac{\text{exp\{} \tau \cdot \kappa(\e_r^i, \e_t^i)\}}{\sum^B_{j=1} \text{exp\{} \tau \cdot \kappa(\e_r^i, \e_t^j)\}},
    \label{eq:loss}
\end{equation}

Here, $\e_r^i$, $\e_t^i$ denote the embeddings of the reformulation encoder and the target encoder for the $i$-th triplet, $\kappa(\cdot, \cdot)$ denotes the cosine similarity, $\tau>0$ is a temperature parameter that controls the range of the logits, and $B$ is the number of triplets in a batch. Both text encoder and the reformulated image encoder are updated during the alignment training.

\input{table/zeroshotmain}
Given reformulated embeddings, we further boost the performance by fusing the pure text embedding with the reformulated embedding using the following settings from most of the previous papers~\cite{Baldrati_2022_CVPR_clip4cir0, liu2023bi, liu2023candidate}.
Given the reformulated embedding $\e_r$, we calculate the final embedding as follows:
\begin{equation}
    \e_f = \lambda \e_r + (1-\lambda) \phi_t(t_{ab}),
\end{equation}
where $\lambda$ is the trained hyper-parameter weight to combine original features from two sides. 
After the whole network is trained, this combiner weight is fine-tuned using the same $\mathcal{L}_{\text{contr}}$ by fixing backbone weights and replacing $\e_r$ with $\e_f$ in Equation~\ref{eq:loss}.

%% file: table/zeroshotmain.tex
\begin{table*}[t]

\centering
\caption{Quantitative comparison with state-of-the-art zero-shot methods (and with supervised baselines). Here, \(^{\star}\) indicates data are sampled from subsets of the Fashion200k and NLVR datasets. For zero-shot methods, evaluation metrics are from the same model across two datasets. For supervised methods, evaluations are from the best models respectively on each dataset. $^{\dagger}$ denotes the training scale is additional to the original CLIP pre-trained model. Candidate R$_{50/100}$ is marked \textcolor{gray}{grey} as it requires a much higher computational cost. \textbf{Results indicate that our InstructCIR reaches state-of-the-art performance on both zero-shot and supervised 
benchmarks. We select the best performances in existing methods for fair comparison. The first section results (image or text only) are with the same CLIP-L14 architecture as our method.  }
\label{tb:zero-shotmain}
} 
\label{table:compare_sota}
\resizebox{1.0\textwidth}{!}{%
\begin{tabularx}{1.08\textwidth}{l c l l c c c c c c c}
\toprule
 & && \multirow{2}{*}[-5pt]{\shortstack{\#Sample\\Scale}}&
 \multicolumn{4}{c}{\textbf{CIRR}}& 
 \multicolumn{3}{c}{\textbf{FashionIQ}} \\
 \cmidrule(lr){5-8} \cmidrule(lr){9-11}
 \multicolumn{1}{c}{Methods}&Zero-Shot& Data Source&  & R@1& R@5& R@50&\text{$\rm R_{Subset}@1$}&R@10& R@50&Average\\
\midrule 
Image Only + CLIP-L14 &\checkmark& CLIP~\cite{radford2021learning} & - &8.42 &23.81&61.03 &22.98 & 6.33  &15.21 & 10.78\\
Text Only + CLIP-L14 &\checkmark& CLIP~\cite{radford2021learning} & - &22.98 &46.83&82.36 &63.88 &19.05  &37.82 & 28.63\\
Image-Text Sum. +CLIP-L14  &\checkmark& CLIP~\cite{radford2021learning} & - &11.71 &35.06& 77.49 &32.77 &24.60  &43.21 & 33.91\\
InstructPix2Pix +CLIP-L14  &\checkmark&  CLIP~\cite{radford2021learning}/ LAION~\cite{schuhmann2021laion}& - &22.03 &47.81&83.52 &61.67 & 9.86  &19.63 & 15.03\\
\midrule
\multicolumn{11}{c}{Zero Shot Benchmarks} \\
\midrule
PALAVRA~\cite{eccv2022_palavra_cohen} &\checkmark& PerVL~\cite{eccv2022_palavra_cohen} & $\sim$1m. &16.62 &43.49&83.95 &41.61 &19.76  &37.25 & 28.51\\
Pic2Word~\cite{saito2023pic2word} & \checkmark & CC3M~\cite{sharma2018conceptual} & 3m. & 23.90 & 51.70 & 87.80 & - & 24.70 & 43.70 & 34.20\\
SEARLE-XL-OTI~\cite{baldrati2023zero}&{\checkmark}& COCO (CIRCO)~\cite{lin2014microsoft}& 118k.$^{\dagger}$ &24.87 &52.31&88.58 &53.80&27.61 &47.90 &37.76\\
CompoDiff w/T5-XL~\cite{gu2023compodiff} & {\checkmark} & SynthTriplets18m~\cite{sharma2018conceptual} & 18m. &19.37 & 53.81&90.85 & 28.96 &{\textbf{37.36}} & 50.85 & {{44.11}}\\ 
CASE Pre-LaSCo.Ca.~\cite{levy2023data}&{\checkmark}& LaSCo~\cite{levy2023data} &360k.$^{\dagger}$ &35.40 &65.78& {{94.63}} &64.29&- &- &-\\
{TransAgg}& {\checkmark}& LAION~\cite{schuhmann2021laion} &32k$^{\dagger}$ &{{37.87}} &{\underline{68.88}} & {{93.86}}& {\underline{69.79}} & {{34.64}} &{\underline{55.72}} &{\underline{45.18}}\\
{COVR~\cite{ventura2024covr}}& {\checkmark}& WebVid-~\cite{ventura2024covr} & 1.6m$^{\dagger}$ &{\textbf{39.28}} &{{68.22}} & \underline{{94.65}}& {-} & {{27.70}} &{{44.63}} &{{36.15}}\\

\textbf{InstructCIR (Ours)}& {\checkmark}& LAION~\cite{schuhmann2021laion} & 300k$^{\dagger}$ &\underline{{38.56}} &\textbf{{69.21}} & \textbf{{95.21}}& {{68.22}} & {{{36.56}}} &{\textbf{56.33}} &{\textbf{46.89}}\\
\textbf{InstructCIR (Ours)}& {\checkmark}& Fashion200k~\cite{han2017automatic}/ NLVR~\cite{suhr2017corpus}$^{\star}$ & 300k$^{\dagger}$ &\textbf{{39.28}} &\textbf{{69.62}} & \textbf{{95.88}}& \textbf{{69.87}} & {\underline{{37.32}}} &{\textbf{56.84}} &{\textbf{47.08}}\\
\midrule
\multicolumn{11}{c}{Compared to Supervised Learning.} \\
\midrule
CLIP4CIR~\cite{baldrati2022conditioned} & $\times$& CIRR/FashionIQ &-&38.53 &69.98&95.93&68.19&38.32&61.74&50.03\\
BLIP4CIR+Bi~\cite{liu2023bi} & $\times$& CIRR/FashionIQ&-&40.15&73.08&96.27&72.10 &43.49 &67.31&55.40\\
CASE~\small{Pre-LaSCo.Ca.$^\dagger$}~\cite{levy2023data} & $\times$& CIRR/FashionIQ &-&{49.35} &{80.02}&{97.47}& \textbf{76.48} &{48.79}& {70.68} & {59.74}\\
Candidate $\mathbf{F}$ ~\cite{liu2023candidate} & $\times$& CIRR/FashionIQ &-&{44.70} & {76.59}&{97.18}& {75.02} & {46.15}& {69.15} &{57.65}\\
\textcolor{gray}{Candidate \small{$\mathbf{R}_{50/100}$}}~\cite{liu2023candidate} & \textcolor{gray}{$\times$}& \textcolor{gray}{CIRR/FashionIQ} &-&\textcolor{gray}{50.55} & \textcolor{gray}{\textbf{81.75}} &\textcolor{gray}{97.18} & \textcolor{gray}{\textbf{80.04}} & \textcolor{gray}{\textbf{51.17}} & \textcolor{gray}{\textbf{73.13}} & \textcolor{gray}{\textbf{62.15}}\\
{COVR~\cite{ventura2024covr}}& $\times$ & CIRR/FashionIQ & - &\underline{{50.41}} & \underline{{80.96}} & \underline{{97.64}}&{{-}} & \textbf{49.40} &{\textbf{70.98}} &{\textbf{60.19}}\\
\textbf{InstructCIR (Ours)}& $\times$ & CIRR/FashionIQ & - &\textbf{{50.70}} & \textbf{{81.61}} & \textbf{{98.27}}&{\underline{76.10}} & \underline{49.03} &{\underline{70.96}} &{\underline{60.00}}\\
\bottomrule
\end{tabularx}
}
\vspace{-1em}
\end{table*}

%% file: sec/4_relatedworks.tex
\section{Related Work}
\paragraph{Composed Image Retrieval (CIR):}
Composed image retrieval (CIR) retrieves images using a reference image-text pair~\cite{vo2019composing,duan2022position,duan2023cross}, with applications in fashion~\cite{wu2021fashion} and scene composition~\cite{liu2021image}. Traditional methods merge latent embeddings~\cite{li2022exploring,li2023clip,li2025closer} from both modalities to form retrieval queries. Techniques range from TIRG's gating and residual connections~\cite{vo2019composing} to VAL's transformer-based hierarchical design~\cite{chen2020image_val}. Wu~\etal\cite{wu2021fashion} employ a custom transformer for early image-language fusion. In contrast, Goenka~\etal\cite{goenka2022fashionvlp} use BERT~\cite{devlin2018bert} for image-text-tag unified coding, while Han~\etal\cite{han2022fashionvil} pre-train a model using a vast fashion dataset. Modern CIR approaches, like CLIP4CIR~\cite{baldrati2022effective} and BLIP4CIR~\cite{liu2023bi}, leverage pre-trained visual-language models and apply late fusion. CASE~\cite{levy2023data} enhances this by adding external data. Candidate-R~\cite{liu2023candidate}, aiming for peak performance, re-ranks retrieval candidates, albeit at a much higher computational cost. Notably, all these strategies require source-prompt-target triplets for training, and the high cost of obtaining such data constrains CIR's broader application.

\paragraph{Zero-Shot Compsed Image Retrieval:}
%
The concept of zero-shot composed image retrieval has recently garnered significant attention. Two contemporaneous studies, Pic2Word~\cite{saito2023pic2word} and ZS-CIR~\cite{baldrati2023zero}, utilize image-caption datasets to train networks that represent images as singular tokens, thus facilitating cross-modal retrieval in the text domain. CompoDiff~\cite{gu2023compodiff} leverages a modified diffusion-denoising model to iteratively refine search queries and introduces a new dataset, SynthTriplet18M. This dataset comprises images synthesized through the prompt-to-prompt model~\cite{hertz2022prompt}, guided by corresponding captions.

Our concurrent work, TransAgg~\cite{liu2023zero}, harnesses ChatGPT combined with human-translated templates on selected image caption data from LAION~\cite{schuhmann2022laion}, yielding impressive results. Distinctly, our approach aims to achieve zero-shot image retrieval relying solely on image distribution. By integrating image captioning models with large language models (LLMs) and capitalizing on scaling potential, InstructCIR sets new benchmarks in the realm of composed image retrieval.

%% file: sec/3_exp.tex
\section{Experiments}
In this section, we provide comprehensive experiments to illustrate the state-of-the-art performance of InstructCIR on both zero-shot and supervised composed image retrieval. 


\subsection{Setup}
 The reformulation network is modified from CLIP pretrained ViT-L/14 and injects cross attention to each layer. The cross-attention layer is initialized with Xavier~\cite{kumar2017weight} initialization.
The cross-attention layer has the same heads as the main backbone. 
The training data is sampled from Fashion200k~\cite{han2017automatic_fashion200k} and NLVR~\cite{suhr2017corpus} which respectively have 280k and  21.4k images. 
For the Fashion200k dataset, we randomly sample image pairs under the same meta class~\footnote{Fashion200k has five meta classes: dresses, jackets, pants, skirts, tops.}, while for NLVR we sample the whole dataset. 
For the image caption model, we directly used LLaVA Vicuna 13B pre-train weights. 
We use our own language model with 33B as the text reformulator, also, we provide a comparison with different language models in supplementary. 

Since the unique combination is scalable, we creating image pairs from 16k up to 500k to report the performance curve. 
The model is trained with AdamW~\cite{loshchilov2017decoupled} optimizer, with learning rate $2 \times 10^{-6}$, weight decay 0.1, batch size 32. 
The model is implemented using PyTorch and trained with eight A100 GPU instances. 

\input{table/circo}

\subsection{Zero-Shot Quantitative Evaluation}
\paragraph{Baselines:}  
To illustrate the efficiency of our proposal, we provide a comparison with a wide range of zero-shot CIR baselines. 
CLIP~\cite{radford2021learning} and PALAVRA~\cite{eccv2022_palavra_cohen} provide baselines that utilize frozen vision-language pretraining models. 
We respectively evaluate \textit{text only}, \textit{image only}, and direct embedding summation (\textit{Image-Text Sum.}) to report the performance of each modality. 
Pic2Word~\cite{saito2023pic2word} and SEARLE~\cite{baldrati2023zero} represent realizing zero-shot CIR by converting an image into a single text token.
Moreover, we provide an image editing baseline by using InstructPix2Pix~\cite{brooks2022instructpix2pix} to edit reference images towards the target image with text prompt then applying pure image retrieval.
CompoDiff~\cite{gu2023compodiff} represents a diffusion-based generative model on latent space. 
CASE~\cite{levy2023data} and TransAGG~\cite{liu2023zero} suggest using LLMs to generate reformulation data but rely on image-text pairs. 

\paragraph{Performance:}
The zero-shot evaluation is conducted across two datasets with the same model weights, FashionIQ~\cite{wu2021fashion} (fashion) and CIRR~\cite{liu2021image} (real-life scenarios) in Table~\ref{tb:zero-shotmain}. 

The first sector provides an intuitive understanding of zero-shot performance by using raw model analysis. Text-only and image-text summation using CLIP~\cite{radford2021learning} exceed the performance of early baseline PALAVRA~\cite{eccv2022_palavra_cohen} and image only.
For recent works, Pic2word~\cite{saito2023pic2word} and SEARLE~\cite{baldrati2023zero} introduce $2.09\%$ to $7.09\%$ performance improvement on recall metrics. 
CompoDiff~\cite{gu2023compodiff} reaches highest $37.36\%$ top 10 recall ($R@10$) on FashionIQ while failing to reach competitive performance on CIRR.
TransAgg~\cite{liu2023zero} and CASE further boost performance by introducing pretraining data. 

While just given image distribution, InstructCIR reaches recall $38.18\%$, $69.62\%$, and $95.88\%$ respectively on $R@\{1,5,50\}$ on CIRR dataset. Using the same model, InstructCIR reaches recall $36.91\%$ and $55.84\%$ respectively on $R@10$ and $R@50$ on the FashionIQ dataset.
\textit{Results demonstrate that our InstructCIR reaches new state-of-the-art performance across two major datasets}.

\paragraph{Extensive Zero-Shot Performance on Benchmark CIRCO}\label{subsec:circo}
We have now compared our method (InstructCIR) with two other zero-shot methods on an extensive dataset proposed recently called CIRCO~\cite{baldrati2023zero}. The observation is similar to other datasets, that our approach clearly outperforms previous methods even on the CIRCO dataset.
We will include these results in the revised manuscript.


\input{table/ablation}

\subsection{Supervised Quantitative Evaluation} \label{subsec:supervised}

To underscore the efficiency of our proposed embedding reformulation network, we benchmarked it using standard supervised learning on two prominent CIR datasets: FashionIQ~\cite{wu2021fashion} (fashion-centric) and CIRR~\cite{liu2021image} (reflecting real-life scenarios). Unlike the zero-shot setting, which evaluates a single model across both datasets, the supervised benchmark trains optimized models for each dataset before evaluation. 

It's worth noting that the Candidate Re-ranking process, which refines results from the top 100 retrieved candidates, incurs additional computational costs. Hence, we've highlighted Candidate \( \mathbf{R}_{100} \) in grey. In the CIRR evaluation, our InstructCIR outperforms most preceding benchmarks, even surpassing the Candidate \( \mathbf{R}_{100} \) in the Recall@K metrics. For the Recall$_{\text{Subset}}$@K metric, while Candidate \( \mathbf{R}_{100} \) achieves the peak performance, both CASE and our proposal are closely competitive for the second-best score. In the FashionIQ evaluation, InstructCIR ranks slightly below CASE, with Candidate \( \mathbf{R}_{100} \) securing the third spot. These results suggest that our network architecture is on par with the state-of-the-art when compared against supervised baselines, underscoring the efficacy of our model design.

\subsection{Ablation Study}
TransAgg~\cite{liu2023zero} and our model significantly outperform previous models as these two models both utilize LLMs to create reformulated text prompts. 
The difference is that TransAgg utilizes existing image caption dataset, but InstructCIR creates captions given the image distribution. 
We further conducted detailed ablation studies in Table~\ref{tb:mainablation} to help people understand the performance improvement.

\paragraph{Architectural Efficiency:}
In this section, we provide an ablation study on architecture efficiency compared to concurrent work TransAgg~\cite{liu2023zero}. 
The first sector of Table~\ref{tb:mainablation} compares TransAgg~\cite{liu2023zero} with our model by exchanging training data.
By controlling data scale the same (32k), TransAgg R@\{1, 5\} of CIRR dataset respectively increases from $32.67\%, 64.05\%$ to $33.26\%, 65.67\%$. This suggest the efficiency of zero-shot data creating pipeline of InstructCIR. 
Also, while using the TransAgg data training, our embedding reformulation network also could reach $35.58\%, 67.85\%$ on CIRR R@\{1, 5\}, which is still higher than previous model. 
This observation suggests that, while using the same data with the same scale of model, our methods outperforms TransAgg clearly. 
The improvement is compared smaller on the FashionIQ dataset, but the conclusion still stands. 

\input{table/modelablation}

Table~\ref{tb:modelablation} provides an ablation study on \textbf{architectual design}. The model is trained with InstructCIR data with a data scale of 32k samples.
When removing EMA~\cite{assran2023self} update, the CIRR R@1 and FashionIQ R@10 drop slightly from $35.83\%$ to $35.04\%$ and from $33.78\%$ to $33.16\%$.  
When further removing both EMA and cross attention to perform embedding reformulation, which is the original CLIP model, the average recall drops significantly by $6.59\% \sim 8.52\%$ on both CIRR and Fashion IQ.
This suggests the design efficiency of the proposed embedding reformulation network. 
We conducted a supervised CIR comparison in Section~\ref{subsec:supervised}, which further illustrates the architectural efficiency by comparing with previous methods with the same training data.

\input{table/retrieval_results}

\paragraph{Language Model Reformulation:}
To illustrate how performance improvements related to pure image distribution we sampled from Fashion200k and NLVR. 
We directly take the images from the supervised dataset, FashionIQ, and CIRR and abandon the annotations. 
The images are fed through the zero-shot pipeline to train the model.
Results in Table~\ref{tb:mainablation} suggest that, although these two datasets show a slightly better performance on data (3.6k and 5.9k) closer to their own distribution, the differences are not significant and they still fall short of models with larger data scales. 
This suggests the good salable property of InstructCIR.

\paragraph{Scaling-Up:} 
The lower part of Table~\ref{tb:mainablation} reports the {scaling-up} experiment of InstructCIR. 
The performance rises from $35.83\%$ R@1 to $38.86\%$ R@1 on the CIRR dataset when scaling from $32$k to $300$k. The findings indicate good scalability, demonstrating that as the data scale increases, so too does performance. 

Furthermore, to obtain an approximate empirical performance upper bound, we augment our approach with semi-supervised data. By utilizing triplet data from the CIRR~\cite{liu2021image} and the FashionIQ~\cite{wu2021fashion} datasets, we further boost the language model by leaking information from the ground truth. 
This exploration is critical, as it provides a trajectory of performance enhancement: starting from a zero-shot scenario and progressively approaching supervised benchmarks.
With the help of the boosted language model, the performance could further rise to $41.32\%$ R@1 on the CIRR dataset and $38.92\%$ R@10 on the FashionIQ dataset, indicated as 33B$^{\mp}$-GT in Table~\ref{tb:mainablation}. This performance \textbf{is even comparable with advanced supervised baselines} CLIP4CIR~\cite{Baldrati_2022_CVPR_clip4cir1} and BLIP4CIR~\cite{liu2023bi}. 

\input{table/testablation}
\paragraph{Test Domain Variance:}
InstructCIR employs LLMs to generate text instructions in a zero-shot manner for training purposes. To showcase the generalization ability of InstructCIR in real-world test distributions, we have conducted an ablation study, presented in Table~\ref{tb:testdomain}. In this study, while retaining the images from the ground truth (GT), we replaced the text GT with generated texts. Our hypothesis was that by using generated texts closer to the training distribution, the model would exhibit improved performance.
Indeed, this change led to an increase in top-1 recall (R@1) as seen in Table~\ref{tb:testdomain}. However, only marginal differences were observed for R@5, R@10, and R@50 metrics. Such results underscore the robust generalization capabilities of our proposed data creation pipeline.

\begin{figure}[t]
    \centering
    \begin{adjustbox}{margin=-0.85cm 0cm 0cm 0cm}
    \includegraphics[width=0.56\textwidth]{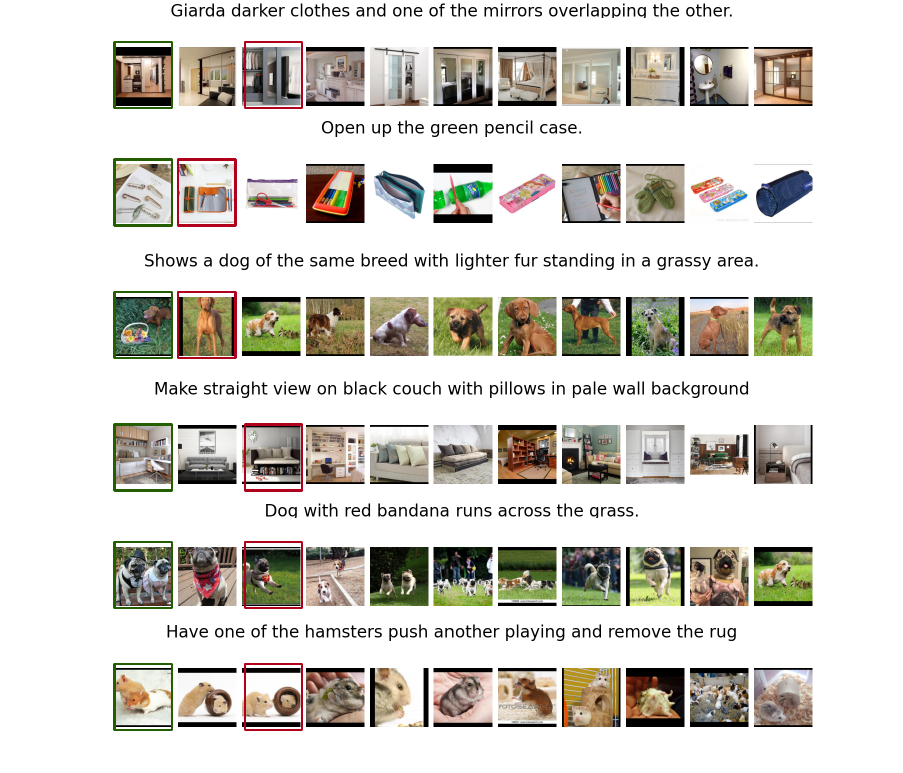}
    \end{adjustbox}
    \caption{Qualitative results on CIRR dataset. \textcolor{greengraff}{green boxes \( \square \)} denote reference images and \textcolor{redgraff}{red boxes \( \square \)} denotes GT label. \label{fig:visualizeCIRR}}
\end{figure}

\subsection{Qualitative Analysis}
To provide a clear understanding of the retrieval performance, we present visualizations of the retrieval results on the FashionIQ dataset (Figure~\ref{fig:visualize}) and CIRR dataset (Figure~\ref{fig:visualizeCIRR}. For each reference image (highlighted with green boxes) paired with text prompts (displayed as the title of each line), we showcase the top 10 retrieved candidates. The target image's ground truth is marked with red boxes on each line.

It's widely recognized that metric-based evaluations can sometimes diverge from human judgments. This is because there may be multiple reasonable results beyond just the ground truths. For instance, in the middle-right line of our visualization, all retrieved results adhere to the prompt ``Has floral design and has flowers and is blue''. However, the ground truth only indicates one of these valid candidates. Common failures occur when the original reference images still appear among the top 10 retrieved candidates. This is especially prevalent when the reference image captures comprehensive human attributes, like a face.

%% file: table/circo.tex
\begin{table}[htbp]
\centering
\vspace{-2pt}
\caption{Comparison on CIRCO dataset~\label{tb:circo}}
\resizebox{0.95\columnwidth}{!}{%
\begin{tabular}{lccccc}
\toprule
Backbone & Method & K = 5 & K = 10 & K = 25 & K = 50 \\
\midrule
\multirow{3}{*}{B/32} & SEARLE-OTI & 7.14 & 7.83 & 8.99 & 9.60 \\
& SEARLE & 9.35 & 9.94 & 11.13 & 11.84 \\
& InstrucCIR & \textbf{10.23} & \textbf{10.98} & \textbf{12.07} & \textbf{13.88} \\
\midrule
\multirow{4}{*}{L/14} & Pic2Word & 8.72 & 9.51 & 10.64 & 11.29 \\
& SEARLE-XL-OTI & 10.18 & 11.03 & 12.72 & 13.67 \\
& SEARLE-XL & 11.68 & 12.73 & 14.33 & 15.12 \\
& InstructCIR & \textbf{12.94} & \textbf{13.84} & \textbf{15.62} & \textbf{16.34} \\
\bottomrule
\end{tabular}
}
\vspace{-2pt}
\end{table}

%% file: table/ablation.tex
\begin{table*}[hbpt]
\centering
\caption{Ablation study on different models, reformulation LLMs, and data distribution with zero-shot setting. 33B$^{\mp}$ denotes our own language model with 33 billion parameters. CIRR$^{\star}$ and FashionIQ$^{\star}$ denotes just utilizing images from GT dataset without annotation but creating zero-shot reformulation using our pipeline. 33B$^{\mp}$-GT denotes we use supervise data to fine-tune the language model. 
\textbf{InstructCIR outperforms TransAgg with respective to the quality of generated data as well as the model architecture.}
\label{tb:mainablation}}
\resizebox{0.98\textwidth}{!}{
\begin{tabularx}{0.95\textwidth}{l llllllllll}
\toprule
&&&&\multirow{2}{*}[-5pt]{\shortstack{\#Reform\\LLM}} && \multicolumn{3}{c}{\textbf{CIRR}} & \multicolumn{2}{c}{\textbf{FashionIQ}}
\\
Model       & Backbone & Images & Caption &  & \#Scale & R@1& R@5&\text{$\rm R_{Subset}@1$}&R@10& R@50 \\
\midrule 
\multirow{3}{*}{TransAgg} & CLIP \small{L/14} & TransAgg   & LAION     & Temp. & 32k & 33.04 & 64.39 & 63.37 & 32.63 & 53.65 \\
 & CLIP \small{L/14} & TransAgg   & LAION     & ChatGPT & 32k & 32.67 & 64.05 & 62.98 & 32.45 & 53.15 \\
 & CLIP \small{L/14} & Fashion/NVLR   & LLaVa     & 33B$^{\mp}$ & 32k & 33.26  & 65.67 & 64.05 & 32.91 & 53.41 \\
                          \hdashline
\multirow{2}{*}{InstructCIR }            & CLIP \small{L/14}& TransAgg      & LAION    & 33B$^{\mp}$ & 32k & 35.58 &	67.85 & 66.87 & 33.52 & 54.07
 \\
                      & CLIP \small{L/14}& Fashion/NVLR      & LLaVa    &33B$^{\mp}$ & 32k & \textbf{35.83}&	\textbf{68.04}&		\textbf{66.93}& \textbf{33.78} & \textbf{54.82}
                      
\\ \midrule 
\multicolumn{11}{c}{Scaling-Up Experiments with Zero-Shot Pipeline} \\ \midrule
 \multirow{7}{*}{InstructCIR }    & CLIP \small{L/14}& CIRR$^{\star}$      & LLaVa     & 33B$^{\mp}$ & 3.6k & 34.74&	65.83& 66.43 & 32.21 & 53.19\\
                    & CLIP \small{L/14}& FashionIQ$^{\star}$      & LLaVa     & 33B$^{\mp}$ & 5.9k & 33.08&	65.98& 66.38 & 33.64 & 54.62 \\
                    & CLIP \small{L/14}& Fashion/NVLR      & LLaVa     & 33B$^{\mp}$ & 32k &35.83&	68.04&		66.93& 33.78 & 54.82\\
                    & CLIP \small{L/14}& Fashion/NVLR      & LLaVa     & 33B$^{\mp}$ & 65k &36.72&68.49&67.03& 34.85 & 55.16 \\
                    & CLIP \small{L/14}& Fashion/NVLR      & LLaVa     & 33B$^{\mp}$ & 95k &36.92 &68.64	&68.34 & 35.94 & 55.78 \\
                    & CLIP \small{L/14}& Fashion/NVLR      & LLaVa     & 33B$^{\mp}$ & 200k &38.04&69.56&69.45  & 36.64 & 56.18\\
                    & CLIP \small{L/14}& Fashion/NVLR      & LLaVa     & 33B$^{\mp}$ & 300k &38.86 &69.62& 69.87 &37.32 & 56.84\\ \cdashline{2-11}
                    & CLIP \small{L/14}& Fashion/NVLR      & LLaVa     & 33B$^{\mp}$-GT & 300k &\textbf{41.32} &\textbf{72.55}& \textbf{71.01} &\textbf{38.92} & \textbf{58.43}\\ 
\bottomrule
\end{tabularx}
}
\vspace{-2ex}
\end{table*}

%% file: table/modelablation.tex
\begin{table}[t]
\centering
\caption{Ablation study on architectural design across CIRR and FashionIQ with zero-shot setting. Cross-attention-based latent fusion yields the most benefit. \label{tb:modelablation}  }
\resizebox{0.5\textwidth}{!}{%
\begin{tabularx}{0.5\textwidth}{llllll} \toprule
                     &       & \multicolumn{2}{l}{\textbf{CIRR}} & \multicolumn{2}{l}{\textbf{FashionIQ}} \\
Model                & \#Scale & R@1         & R@5        & R@10          & R@50          \\ \midrule
InstructCIR          & 32k   & 35.83       & 68.04      & 33.78         & 54.82         \\
+w/o EMA              & 32k   & 35.04       & 67.85      & 33.16         & 54.23         \\
\multirow{1}{*}{\shortstack{+w/o CrossAtt.} } & 32k   & 28.45       & 59.33      & 24.53         & 45.67   \\
\bottomrule
\end{tabularx}
}
\end{table}

%% file: table/retrieval_results.tex
\begin{figure*}[t]
    \centering
    \begin{adjustbox}{margin=-0.75cm 0cm 0cm 0cm}
    \includegraphics[width=1.05\textwidth]{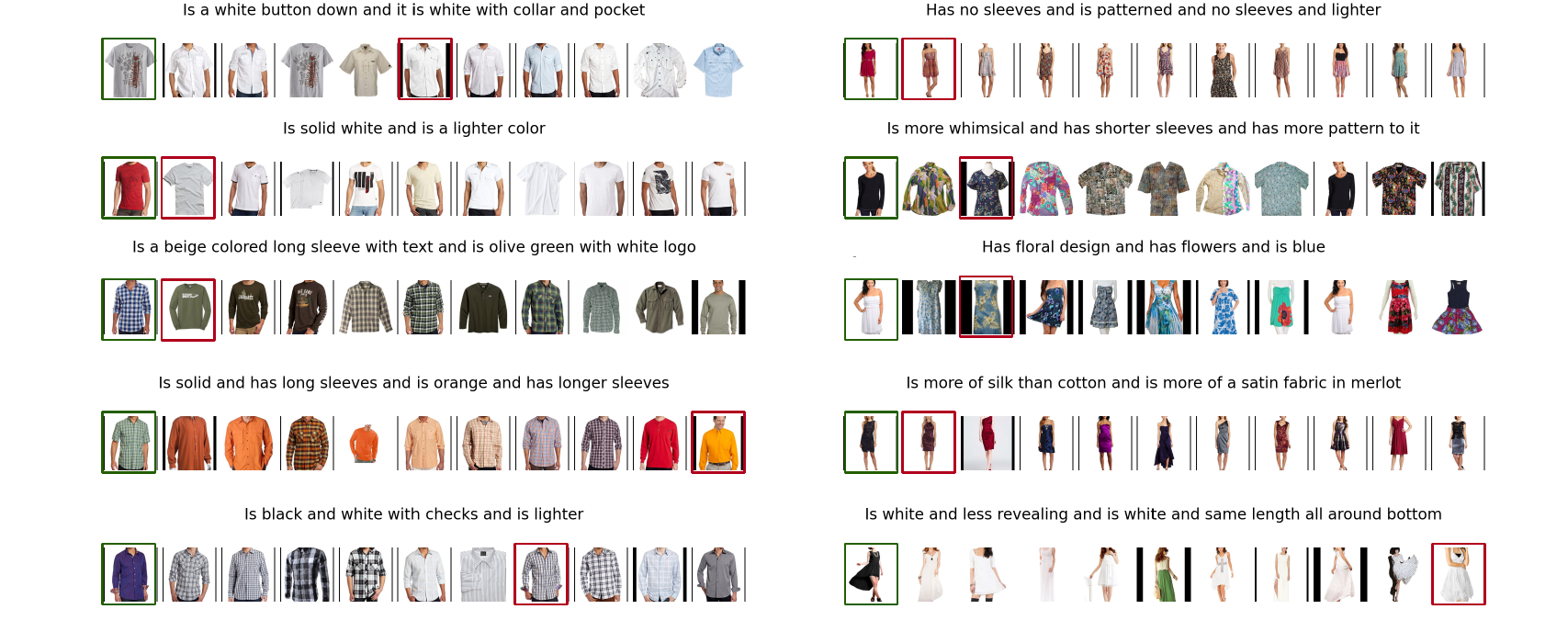}
    \end{adjustbox}
    \caption{Qualitative results on FashionIQ dataset. For each given reference image and text instructions, we visualize the top 10 retrieved candidates. Here, \textcolor{greengraff}{green boxes \( \square \)} denote reference images, and \textcolor{redgraff}{red boxes \( \square \)} denotes target images indicated by ground truth label. \label{fig:visualize}}
\end{figure*}

%% file: table/testablation.tex
\begin{table}[hbpt]
\centering
\caption{Ablation study by replacing GT text instructions by generated text instructions.  Small difference indicate the domain gap between generated and GT reformulation texts is small. \label{tb:testdomain}  }
\resizebox{0.45\textwidth}{!}{%
\begin{tabularx}{0.45\textwidth}{llllll} \toprule
                     &       & \multicolumn{2}{l}{\textbf{CIRR}} & \multicolumn{2}{l}{\textbf{FashionIQ}} \\
Image                & Test Text & R@1         & R@5        & R@10          & R@50          \\ \midrule
GT       & GT   &   35.83    &   68.04     &  33.78        &   54.82      \\
GT & Generated   &   37.45      &  68.31    &    33.92      &  54.87 \\ \hdashline
\multicolumn{2}{c}{$\Delta$} &   +1.62      &  +0.27    &   +0.14    &  +0.04 \\ 
\bottomrule
\end{tabularx}
}
\end{table}

%% file: sec/5_conclusion.tex
\section{Limitations}
Achieving optimal results requires sampling image pairs that have complementary features but maintain certain shared characteristics. When images with excessive dissimilarity are chosen, LLMs tend to falter, often generating descriptions that encompass both images rather than high-quality reformulation text prompts. However, such extreme scenarios are uncommon in CIR, where reference and target images usually differ only at a fine-grained level. Experiments show that these dissimilarities are tolerable as the training scale increases, without causing significant negative impacts on the model.
\section{Conclusion}
This work explores achieving zero-shot composed image retrieval based solely on an unannotated image collection. Our approach involves transforming images into detailed captions and then generating reformulated text prompts within the text domain. With the support of advanced large language models, this data-creating approach is not only efficient but also scalable, without requiring pre-existing caption data. 
Leveraging this scalable pipeline, InstructCIR reaches new state-of-the-art in the zero-shot domain. Furthermore, our proposed embedding reformulation network also attains state-of-the-art results in supervised benchmarks, underscoring the efficacy of our design. 